\documentclass[graybox]{svmult}
\usepackage{mathptmx}       
\usepackage{helvet}             
\usepackage{courier}            
\usepackage{type1cm}         

\usepackage{makeidx}         
\usepackage{graphicx}         
\usepackage{multicol}          
\usepackage[bottom]{footmisc} 

\makeindex             

\usepackage{graphicx}
\usepackage{natbib}
\usepackage[utf8x]{inputenc}

\newcommand{\cf}{cf.~}
\newcommand{\eg}{e.g.,~}
\newcommand{\ie}{i.e.,~}
\newcommand{\code}[1]{{\texttt{\small #1}}}
\newcommand{\reftab}[1]{{Table~\ref{#1}}}
\newcommand{\reffig}[1]{{Figure~\ref{#1}}}
\newcommand{\refsec}[1]{{Section~\ref{#1}}}

\newcommand{\mte}{MULTEXT-East}
\newcommand{\mt}{MULTEXT}

\newcommand{\eagles}{EAGLES}

\begin{document}

\title*{MULTEXT-East}
\author{Tomaž Erjavec}
\institute{Tomaž Erjavec \at
              Dept. of Knowledge Technologies, 
		Jožef Stefan Institute \\
		Jamova cesta 39 \\
		SI-1000 Ljubljana \\
		Slovenia\\
        \email{tomaz.erjavec@ijs.si}
}

%
%
\maketitle

\abstract{The chapter presents the
  MULTEXT-East language resources, a multilingual dataset for language
  engineering research, focused on the morphosyntactic level of linguistic description.  
  The MULTEXT-East dataset includes the EAGLES-based morphosyntactic
  specifications, morphosyntactic lexicons, and an annotated multilingual corpora. 
  The parallel corpus, the
  novel ``1984'' by George Orwell, is sentence aligned and contains hand-validated
  morphosyntactic descriptions and lemmas.
  The resources are uniformly encoded in XML, using the Text Encoding Initiative
  Guidelines, TEI P5, 
  and cover 16 languages: 
  Bulgarian, 
  Croatian, 
  Czech, 
  English, 
  Estonian,
  Hungarian,
  Macedonian, 
  Persian,  
  Polish, 
  Resian, 
  Romanian, 
  Russian, 
  Serbian,
  Slovak, 
  Slovene, and
  Ukrainian.
  This dataset is
  extensively documented, and freely available for research purposes.
  This case study gives a history of the development of the MULTEXT-East resources, 
  presents their encoding and components, 
  discusses related work and gives some conclusions.}



\section{Introduction}

The \mte\  project, (Multilingual Text Tools and Corpora for Central and Eastern European Languages) ran from ’95 to ’97 and developed standardised language resources for six Central and Eastern European languages, as well as for English, the ``hub'' language of the project \citep{coling:mte}. The project was a spin-off of the MULTEXT project \citep{mt94}, which pursued similar goals for six Western European languages.  
The main results of the project were morphosyntactic specifications, defining the tagsets for lexical and corpus annotations in a common format, lexical resources and annotated multilingual corpora. In addition to delivering resources, a focus of the project was also the adoption and promotion of encoding standardization. On the one hand, the morphosyntactic annotations and lexicons were developed in the formalism used in MULTEXT, itself based on the specifications of the Expert Advisory Group on Language Engineering Standards, EAGLES \citep{eagles}. 
On the other, the corpus resources were encoded in SGML, using CES, the Corpus Encoding Standard \citep{lrec:ces}, a derivative of the Text Encoding Initiative Guidelines, version P3, \citep{TEI94}.

After the completion of the EU MULTEXT-East project a number of further projects have helped to keep the MULTEXT-East resources up to date as regards encoding and enabled the addition of new languages. The latest release of the resources is Version 4 \citep{lrec10-mte,ijlre:mtev4}, which covers 16 languages. The main improvements to Version 3 were the addition of resources for five new languages, updating of four, and the recoding of the morphosyntactic specifications from \LaTeX\ to XML:
the specifications and the corpora are now uniformly encoded to a schema based on the latest version of the Text Encoding Initiative Guidelines, TEI P5 \citep{TEIP5}.

The resources are freely available for research and include uniformly encoded basic language resources for a large number of languages. 
These mostly include languages for which resources are scarcer than those for English and the languages of Western Europe. 
Best covered are the Slavic languages, which are well known for their complex morphosyntax and \mte\ is the first dataset that enables a qualitative and quantitative comparison between them on this level of description.

The MULTEXT-East resources have helped to advance the state-of-the-art in language technologies in a number of areas, e.g., part-of-speech tagging \citep{tufis:lnia,hajic00}, learning of lemmatisation rules \citep{DzeroskiErjavec2004,Toutanova_ACL05},  word alignment \citep{tufis:cheap,Martin05wordalignment}, and word sense disambiguation \citep{ide:crosswsd,IdeErjTuf2002}. 
They have served as the basis on which to develop further language resources, e.g. the WordNets of the BalkaNet project \citep{Balkanet} 
and the JOS linguistically tagged corpus of Slovene \citep{lrec:jos2}. 
The morphosyntactic specifications have  become a de-facto standard for several of the languages, esp.\ Romanian, Slovene and Croatian, where large monolingual reference corpora are using the \mte\ tagsets in their annotation. 
The resources have also provided a model to which some languages still lacking publicly available basic language engineering resources (tagsets, lexicons, annotated corpora) can link to, taking a well-trodden path; in this manner resources for several new languages have been added to the Version 4 resources. 

\reftab{tab:overview} summarises the language resources of MULTEXT-East Version 4 by language (similar languages are grouped together and the ordering roughly West to East), and by resource type. 
The resources marked by X are present in Version 4, while the ones marked with O have been already produced and will be released in the next version. 
The meaning of the columns is the following: 
\begin{itemize}
\item {MSD specs}: morphosyntactic specifications, defining the features and tagsets of morphosyntactic descriptions (MSDs) of the languages;
\item {MSD lexicon}: morphosyntactic lexicons;
\item {1984 MSD}: MSD and lemma annotated parallel corpus, consisting of the novel “1984” by G. Orwell, approx. 100,000 tokens per language;
\item {1984 alignments}:  sentence alignments over the “1984” corpus;
\item {1984 corpus}:  a variant of the parallel corpus, extensively annotated with structural information (\eg paragraph, verse, quoted speech, note, etc.), named-entity information (name, number), and basic linguistic information (foreign, sentence);
\item {Comparable corpus}: multilingual corpus comprising comparable monolingual structurally annotated texts of fiction (100,000 tokens per language) and newspaper articles (also 100,000 tokens per language);
\item {Speech corpus}: parallel speech corpus, 200 sentences per language, spoken + text.
\end{itemize}

\begin{table}
\caption{MULTEXT-East resources by language and resource type.}
\label{tab:overview}
\begin{tabular}{|l|l|c|c|c|c|c|c|c|}
\hline
Language &Language family&\textbf{MSD} &\textbf{MSD} &
\multicolumn{3}{c|}{\textbf{1984}}&Comparable&Speech\\
  &             &\textbf{specs}&\textbf{lexicon}&
\textbf{MSD}&\textbf{alignments}&corpus&corpus&corpus\\
\hline
English & Germanic & X & X & X & X & X & X & - \\
Romanian & Romance & X & X & X & X & X & X & X \\
Polish & West Slavic & X & X & X & O & - & - & - \\
Czech & West Slavic & X & X & X & X & X & X & - \\
Slovak & West Slavic & X & X & X & O & - & - & - \\
Slovene & South West Slavic & X & X & X & X & X & X & X \\
Resian & dialect of Slovene & X & X & - & - & - & - & - \\
Croatian & South West Slavic & X & O & - & - & - & - & - \\
Serbian & South West Slavic & X & X & X & X & X & - & - \\
Russian & East Slavic & X & X & O & O & X & - & - \\
Ukrainian & East Slavic & X & X & - & - & - & - & - \\
Macedonian & South East Slavic & X & X & X & X & - & - & - \\
Bulgarian & South East Slavic & X & X & X & X & X & X & - \\
Persian & Indo-Iranian & X & X & X & - & - & - & - \\
Estonian & Finno-Ugric & X & X & X & X & X & X & X \\
Hungarian & Finno-Ugric & X & X & X & X & X & X & X \\
\hline
\end{tabular}
\end{table}

We discuss only the resources given in bold in the table, giving the ``morphosyntactic triad'' of \mte, consisting of
the specifications, the lexicon and annotated corpus. 
These resources have had the most impact and are also the most interesting from the point of view of encoding and content.
The structurally annotated parallel and comparable corpora and the speech corpus have been retained from the original \mte\ project but are 
too small, esp.\ from today's perspective, to be really useful. 

The rest of this chapter is structured as follows:
Section \ref{sec:tei} introduces the TEI encoding as used in the \mte\ resources,
Section \ref{sec:msd} details the morphosyntactic specifications and lexicons,
Section \ref{sec:1984} the linguistically annotated parallel ``1984'' corpus,
Section \ref{sec:related} discusses related work, and
Section \ref{sec:conc} gives conclusions and directions for further work.

\section{Resource encoding}
\label{sec:tei}

The MULTEXT-East resources, including the morphosyntactic specification, corpora, alignments, as well as supporting documentation, are all encoded to a common schema following 
the Text Encoding Initiative Guidelines, TEI P5 \citep{TEIP5}. 
The first version of the resources was encoded to the Corpus Encoding Standard, CES \citep{lrec:ces}, 
with subsequent versions moving to XCES \citep{lrec:xces}, the XML version of CES, and later on to TEI, as it is more general, 
defines how to introduce extensions to the core standard and has extensive support. 
This TEIfication finished with Version 4, where the last part of the resources, namely the morphosyntactic specifications (previously as a document typeset using \LaTeX) 
and sentence alignments (previously in XCES) were re-coded to TEI P5. 

The advantages of having all the resources in XML are obvious: they can be edited, validated, transformed, and queried with XML tools. 
Using TEI means that much of needed functionality (schema, documentation, some transformations) is already in place.
For example, the TEI provides a sophisticated set of XSLT stylesheets for converting TEI documents into HTML and other formats, 
useful for putting on-line the MULTEXT-East documentation and the morphosyntactic specifications. 

TEI P5 schemas are constructed by writing a TEI customization, \ie a complete (but possibly quite short) TEI document, 
where the text contains a special element, \code{schemaSpec}, giving the schema in the high-level 
TEI ODD language, with the acronym meaning ``One Document Does it all'' \citep[Chp.\ 22]{TEIP5} . 
The schema specification contains invocations of the needed TEI modules, which define their elements and attributes. 
An ODD document can be processed by an ODD processor, which will generate an appropriate XML schema. 
The XML schema can be expressed using any of the standard schema languages, such as 
ISO RELAX NG (REgular LAnguage for XML Next Generation) \citep{RELAXNG} or the W3C Schema language.
These output schemas can then be used by any XML processor such as a validator or editor to validate or otherwise process documents. 
TEI provides an ODD processor called Roma, also available via a Web interface that helps in the process of creating a customized TEI schema. 

The \mte\ XML schema distributed with the resources consists of the  customisation TEI ODD and the Roma generated Relax NG, W3C and DTD schemas, 
as well as customisation-specific documentation.

\section{The morphosyntactic specifications}
\label{sec:msd}

The morphosyntactic specifications define word-level features (most of) which reside on the interface between morphology and syntax. 
So, for example, the features will not specify the paradigm label of a word, such as ``first masculine declension'', which is a purely morphological feature, nor the valency of a verb, which is a syntactic feature and has no reflex in the morphology of a verb. They will, however, give the part-of-speech, and, depending on the language, gender, number, case, etc., which, on the one hand, are marked on the form of a word (typically its ending) and, on the other, enter into syntactic relationships such as agreement.

In addition to defining features (attributes and their values) the specifications also give the mapping from feature-structures used to annotate word-forms to 
morphosyntactic descriptions (MSDs), which are compact strings used in
the morphosyntactic lexicons and for corpus annotation. 
So, for example, the feature-structure 
\texttt{Noun}, \texttt{Type = common}, \texttt{Gender = neuter}, \texttt{Number = dual},
\texttt{Case = locative} maps to the MSD \texttt{Ncndl}. 
The feature structures can thus be viewed as a logical form of the features associated with a word-form, while the corresponding MSDs is its serialisation.
In addition to the formal parts, the specifications also contain commentary, bibliography, etc.

Although the encoding of the specifications has changed substantially, 
their structure still  follows the original \mt\ specifications: they are composed of the introductory part, 
followed by the common specifications, and then by language particular specifications, one for each 
language. 
The remainder of this section explains the structure of the common specifications, 
the language particular sections, the MSD tagset(s) and their relation of feature-structures, 
an overview of the XSLT stylesheets used to process the specifications 
and the morphosyntactic lexicons.

\subsection{Common specifications}
\label{sec:common}

The common part of the specification gives the 14 \mt\ defined categories, 
which mostly correspond to parts-of-speech, with a few introduced for technical reasons. 
Each category has a dedicated table defining its attributes, their values, 
and their mapping to the (common) MSD strings.
For each attribute-value pair it also specifies the languages it is appropriate for. 
Furthermore, attributes or their values can have associated notes.

\begin{table}[htbp]
\centering
\caption{MULTEXT categories with the number of MULTEXT-East defined attributes, values and languages}\label{tab:cats}
\begin{tabular}[c]{l|c|r|r|r}
Category&Code&Attributes&Values&Languages\\ \hline
Noun&N&14&68&16\\
Verb&V&17&74&16\\
Adjective&A&17&79&16\\
Pronoun&P&19&97&16\\
Determiner&D&10&32&3\\
Article&T&6&23&3\\
Adverb&R&7&28&16\\
Adposition&S&4&12&16\\
Conjunction&C&7&21&16\\
Numeral&M&13&81&16\\
Particle&Q&3&17&12\\
Interjection&I&2&4&16\\
Abbreviation&Y&5&35&16\\
Residual&X&1&3&16\\
\end{tabular}
\end{table}

\reftab{tab:cats} lists the defined categories and, for each category, gives the number of attributes it distinguishes, the number of different attribute-value pairs, and the number of \mte\ languages which use the category.
The feature-set is quite extensive, as many of the languages covered have very rich inflection, are typologically quite different (inflectional, agglutinating) but also have different linguistic traditions.

The definitions for each category are encoded simply as a TEI \texttt{table}, with 
\texttt{@role} giving the function of each row and cell. 
Figure \ref{fig:common} gives the definition of the 
Formation attribute belonging to the Particle category. 
The example states that (a Particle) has the attribute Formation with two values,  simple and compound and both values are valid for Bulgarian, Macedonian and Russian. 
Furthermore, in Particle MSDs, such as \texttt{Qzs} or \texttt{Qgc}, the Formation attribute has position \texttt{2} (taking the category position as 0), 
with the code \texttt{s} or \texttt{c}. 

\begin{figure}
\centering
\begin{verbatim}
<row role="attribute">
 <cell role="position">2</cell>
 <cell role="name">Formation</cell>
 <cell>
  <table>
   <row role="value">
    <cell role="name">simple</cell>
    <cell role="code">s</cell>
    <cell role="lang">bg</cell>
    <cell role="lang">mk</cell>
    <cell role="lang">ru</cell>
   </row>
   <row role="value">
    <cell role="name">compound</cell>
    <cell role="code">c</cell>
    <cell role="lang">bg</cell>
    ...
\end{verbatim}
\caption{Example of the encoding for the common tables.}
\label{fig:common}
\end{figure}

\subsection{Specifications for individual languages}
\label{sec:particular}

The second main part of the specifications consists of the language-specific sections. 
These, in addition to the introductory matter, also contain sections for each category with their tables of attribute-value definitions. 
These tables are similar to the common tables in that they repeat the attributes and their values, although only those appropriate for the language. 
However, they can also re-specify the position of the attributes in the MSD string, leading to much shorter and more readable MSD tags for the language. 
So, if an MSD needs to be uniquely interpreted in a multilingual setting, 
then the mapping from the features to the MSD is made using the common tables,
if not, then the language specific mapping can be used.

The tables can also contain localisation information, \ie in addition to English, 
the names of the categories, attributes, their values and codes can be translated into some other language(s).
This enables expressing the feature-structures and MSDs either in English or in the language that is being described, making them much more suitable for use by 
native speakers of the language.
Such localisation information enables \eg the mapping of the already mentioned MSD
\code{Ncndl} to the Slovene 
\code{Sosdm}, which corresponds to
\code{samostalnik vrsta = občno\_ime}, \code{spol = srednji}, \code{število = dvojina}, \code{sklon = mestnik}.
\reffig{fig:particular} shows the language specific table for the Slovene Particle, which has no attributes. 
As in the common tables, the \code{role} attribute gives the meaning of the cell, while the language of the cell is specified by the \code{xml:lang} attribute.

\begin{figure}
\centering
\begin{verbatim}
    <div type="section" select="sl" xml:id="msd.Q-sl">
      <head>Slovene Particle</head>
      <table n="msd.cat" select="sl" xml:id="msd.cat.Q-sl">
        <head>Slovene Specification for Particle</head>
        <row role="type">
          <cell role="position">0</cell>
          <cell role="name"  xml:lang="sl">besedna_vrsta</cell>
          <cell role="value" xml:lang="sl">členek</cell>
          <cell role="code"  xml:lang="sl">L</cell>
          <cell role="name"  xml:lang="en">CATEGORY</cell>
          <cell role="value" xml:lang="en">Particle</cell>
          <cell role="code"  xml:lang="en">Q</cell>
        </row>
      </table>
       ...
\end{verbatim}
\caption{Encoding of language particular tables.}
\label{fig:particular}
\end{figure}

The language particular section can also contain information on the allowed combinations of particular attribute values
in the form of a list of constraints for each category. This mechanism has been carried over from \mt\ although is has not, 
as yet, been operationalized.

Finally, each language particular section contains an index (also encoded as a table) containing all the valid MSDs, 
\ie it specifies the tagset for the language.
This is an important piece of information, as a tagged corpus can then be automatically validated against this authority list, 
and the tagset can be statically transformed into various other formats, \cf \refsec{sec:xslt}. 
\reffig{fig:msdindex} shows one row from the Slovene MSD index. 
The MSD is given in cell (1), while the rest of the row gives explicative information: 
its expansion into features (2), 
the MSD localised to Slovene (3) and its expansion (4), 
the number of word tokens (5) and types (6) tagged with this MSD in a corpus, 
and (7) examples of usage in the form word-form/lemma. 
The latter three pieces of information have been in this instance automatically 
produced from a corpus and supplementary lexicon of Slovene, 
and the examples chosen are the most frequent word forms in the corpus for the MSD.

\begin{figure}
\centering
\begin{verbatim}
<row role="msd">
  <cell role="msd" xml:lang="en">Ncmsg</cell>
  <cell role="verbose" xml:lang="en">Noun Type=common 
      Gender=masculine Number=singular Case=genitive</cell>
  <cell role="msd" xml:lang="sl">Somer</cell>
  <cell role="verbose" xml:lang="sl">samostalnik vrsta=občno_ime 
      spol=moški število=ednina sklon=rodilnik</cell>
  <cell>15945</cell>
  <cell>2649</cell>
  <cell>časa/čas, sveta/svet, denarja/denar, zakona/zakon, 
      sistema/sistem, konca/konec, maja/maj, programa/program, 
      prostora/prostor, odstotka/odstotek</cell>
</row>
\end{verbatim}
\caption{A row from the MSD index for Slovene.}
\label{fig:msdindex}
\end{figure}

In contrast to Slovene, the MSD lists were extracted directly from the corresponding \mte\ lexicon for each language.
The numbers of MSDs of course differ significantly, although not only due to the inherent differences between the languages 
but also because of different approaches taken in the construction of the lexica: 
while, for some languages, the lexica contain the complete inflections paradigms of the included lemmas, 
others include only word-forms (and their MSDs) that have actually been attested in a corpus. 

English, as a poorly inflecting language, has the lowest number of MSDs, namely 135, and even this number is considerably larger than most English tagsets, 
as the \mte\ specifications introduce quite fined grained distinctions. 
Next come languages that either have ``medium-rich'' inflections (Romanian with 615 or Macedonian with 765 MSDs) or 
list only corpus-attested MSDs (Bulgarian with 338 or Estonian with 642 MSDs), 
followed by inflectionally very rich (South) West Slavic languages (Czech with 1,452 or Slovak with 1,612 MSDs). 
By far the largest tagset is that of the agglutinating Hungarian language (17,279 MSDs),
which can pile many different suffixes (and their features) onto one word-form, resulting in a huge theoretically possible MSD tagset. 
This tagset shows the limits of the MSD concept, as it would most likely be impossible to construct a corpus of 
sufficient size to contain training examples covering all the MSDs. 

\subsection{PoS tags, MSDs and features}
\label{sec:theoretic}

The MSDs have a central status in \mte, as they tie together the specifications, lexicon and corpus and this section discusses the relation between traditional Part-of-Speech (PoS) tagsets, MSDs and features in somewhat more detail.

It should first be noted that in \eagles, as in \mt, the MSDs were not meant to be used in corpus annotation.
Rather, the MSDs were to be mapped to PoS tagsets. PoS tagsets, as traditionally conceived for annotating monolingual corpora, such as the English Penn TreeBank tagset or the 
Stuttgart-T\"{u}bingen (SST) tagset (\cf the Chapter on German Treebanks), 
are not analytical, \ie a tag cannot in the general case be decomposed into morphosyntactic features. 
Especially morphosyntactic features which lead to high ambiguity and are difficult to predict by taggers 
are left out of the tags and PoS tags can even be assigned to individual words, such as ``EX'' for ``Existential there''.
But developing an ``optimal'' mapping of MSDs to tagger-friendly tagsets for individual languages is quite difficult and has not been attempted for most languages, at least not in the scope of \mte. 

A MSD tagset itself, in spite of its seeming simplicity, is also difficult to define unambiguously. 
One reason is the varied interpretations of the `-' symbol. The hyphen is used in MSDs to mark the value of a ``non-applicable'' attribute, either for the whole language, for a combination of features or for a particular lexical item. 
For example, Adverbs have Degree marked on the 2nd position, and Clitic on 3rd, so 
\code{Rgpy} is \code{Adverb Type=general Degree=positive Clitic=yes}, but as adverbial Participles do not distinguish Degree, \code{Adverb Type=particle Clitic=yes} 
will be coded as \code{Rp-y}.  
The same logic applies to cases where an inflectional feature is evident only on some forms and not on others. 
For example, Slovene nouns only distinguish animacy if they have the masculine gender and even here only in the singular accusative form, 
so it is marked only on this form, while the others set the value of Animacy to non-applicable. 

The use of hyphens also brings with it the question whether or not to write trailing hyphens up to the number of attributes defined for the category; in \mt\ they were written, but in \mte\ it was decided to omit them, resulting in \code{Rgpy} rather than \code{Rgpy----}.

With the addition of new languages the number of attributes became quite large, and, as the new attributes were added at the end, 
the MSD strings often became very long (\eg \code{Dg--------q}), which is precisely the reason the language-particular orderings of attributes were introduced. 
However, this does give the option of expressing the MSDs for the same feature set in two ways, according to the common tables, or to the language particular ones. 
A better option would most likely have been to move the MSD constructors completely into the language particular tables, as they are really defined on the level of an individual language.
If the need arises for the MSDs of several languages to be mixed in a corpus they would be easy enough to distinguish by, say, prefixing them by the language code or interpreting them in the context of the superordinate \code{@xml:lang} attribute.

Another complication arises from the fact that features are defined for each category separately; attributes with identical names can have different values with different categories, and the position of the attribute in the MSD string is typically also different for different categories.  
This is in contrast with the mapping between attributes and their positions in the MSD tags which are used for annotating many Czech language resources, 
such as the Prague Dependency TreeBank \citep{PDT:V20}, where each attribute has a fixed position, regardless of the category. 
If MSDs are taken to correspond to fully ground feature-structures, then the PDT system is untyped, while in the \mt\ approach 
categories act as types in the sense of \cite{Car:LogTFS}, each introducing its own attributes. 
Alternatively, and more usefully, an attribute can be taken to be defined by its name, and its valid values as the union of all category-dependent values, \ie as in the PDT. 
This is probably uncontroversial for attributes like \code{Gender}, but more problematic for \eg \code{Type}, which mostly has disjoint values for different categories.

As the preceding discussion shows, there are a variety of ways of writing ``the same'' MSD, and to which exact feature-structure to map the MSD. 
The \mte\ distribution provides translation tables between MSDs and several feature structure encodings and 
the list below gives most of the available options, on one example from the Slovene MSD tagset:
\begin{enumerate}
\item MSD with attribute ordering according to the common specifications, in English: \code{Vmn-----------e}
\item same as 1, but with attribute ordering according to the language particular specifications for Slovene: \code{Vmen}
\item same as 2, but in Slovene: \code{Ggdn}
\item minimal feature set, giving only instantiated features, in English: \code{Verb}, \\
\code{Type=main}, \code{Aspect=perfective}, \code{VForm=infinitive}
\item same as 4, but in Slovene: \code{glagol}, \code{vrsta=glavni}, \code{vid=dovršni}, \\
\code{oblika=nedoločnik}
\item Canonical (type-free) feature set, giving all the attributes defined for the language, in English: \code{Verb}, \code{Type=main}, \code{Gender=0}, \code{Number=0}, \code{Case=0}, \code{Animate=0}
\code{Aspect=perfective}, \code{VForm=infinitive}, \code{Person=0}, \code{Negative=0}, \code{Degree=0}, \code{Definiteness=0}, \code{Owner\_Number=0}, \code{Owner\_Gender=0}, \code{Clitic=0}, \code{Form=0}
\item Canonical (type-free) feature set, giving all the attributes defined in \mte, in English: 
\code{Verb}, \code{Type=main}, \code{Gender=0}, \code{Number=0}, \code{Case=0}, \\
\code{Definiteness=0}, \code{Clitic=0}, \code{Animate=0}, \code{Owner\_Number=0}, \code{Owner\_Person=0}, \code{Owned\_Number=0}  \code{Case2=0}, \code{Human=0}, \code{Aspect=perfective}, \code{Negation=0}, \\ \code{VForm=infinitive}, \code{Tense=0}, \code{Person=0}, \code{Voice=0}, \code{Negative=0}, \code{Clitic\_s=0}, \code{Courtesy=0}, \code{Transitive=0}, \code{Degree=0}, \code{Formation=0}, \code{Owner\_Gender=0}, \\ \code{Referent\_Type=0}, \code{Syntactic\_Type=0}, \code{Pronoun\_Form=0}, \code{Wh\_Type=0}, \\ \code{Modific\_Type=0}  \code{Coord\_Type=0}, \code{Sub\_Type=0}, \code{Form=0}, \code{Class=0}
\end{enumerate}

\subsection{XSLT stylesheets}
\label{sec:xslt}

The \mte\ specifications come with associated XSLT stylesheets, which 
take the specifications as input, usually together with certain parameters, 
and produce either XML, HTML or text output, depending on the stylesheet.
Three classes of transformations are provided: 
the first help in adding a new language to the specifications themselves; 
the second transform the specifications into HTML for reading; 
and the third transform (and validate) a list of MSDs. 
The outputs of the second and third class of transformation are included in the \mte\ distribution. 

There are two stylesheets for authoring the specifications for a new language. 
The first stylesheet (\code{msd-split.xsl}) takes the common part of the specifications and, as a parameter, 
a list of languages, and produces the language specific section for a new language, 
copying into it all the features defined for the selected languages. 
The intention is to make it easier to author the language specific specifications for a new language, 
by constructing a template that already contains the features of the language(s) that are most similar to it. 
The second stylesheet (\code{msd-merge.xsl}) takes the language specific section for a new or updated language and the common part, 
and produces the common part with the language added or updated. 
This might involve only adding the language flag to existing attribute-value pairs, 
but also adding or deleting attributes or values from the common tables.
The stylesheet warns of such cases, making it also suited for validating language specific sections against the common tables.

For converting the specifications into HTML the stylesheet \code{msd-spec2prn.xsl} is first used to pre-process them in order to add various 
indexes (of attributes, values, MSDs) and to convert the tables into a more human readable form, 
which largely follows the formatting of the original MULTEXT(-East) specifications. 
This pre-processed version of the specifications is still in TEI XML and 
is then fed through the standard TEI XSLT stylesheets to produce the HTML (or other) output. 

Finally, there are two stylesheets that take the specifications and a list of MSDs and 
converts this list into various other formats. 
The stylesheet \code{msd-expand.xsl} produces different types of output, depending on the values of its parameters. 
It can check an MSD list for well-formedness against the specifications 
or can produce an expansion of the MSDs into their feature structure equivalents. 
Here it distinguishes several expansions, most already presented in the previous section, 
from a brief one, meant to be the shortest human readable expansion, 
to the full canonical form, where all the defined attributes are listed.
The stylesheet can also produce the collating sequence for the MSDs with which it is possible to sort MSDs 
so that their order corresponds to the ordering of categories, attributes and their values in the specifications. 
Finally, the stylesheet is able to localise the MSD or features on the basis of the language specific section with localisation information. 
The second stylesheet, \code{msd-fslib.xsl} transforms the MSD list into TEI feature and feature-structure libraries, 
suitable for inclusion into TEI encoded and MSD annotated corpora. 


\subsection{The morphosyntactic lexicons}
\label{sec:lex}

The \mte\ lexicons contain mid-sized lexicons for most of the languages and
are, from the encoding standpoint, very simple. 
Each lexicon is a tabular file with one entry per line, composed of three fields: 
(1) the \emph{word-form},
which is the inflected form of the word, as it appears in the text,
modulo sentence-initial capitalisation; (2) the \emph{lemma}, 
the base-form of the word, which \eg serves as the head-word in a dictionary; 
and (3) the \emph{MSD}, \ie the
morphosyntactic description, according to the language particular specifications.

It should be noted that the lexicon is necessary to ground the specifications and make them useful:
it is only by associating a MSD with lexical items (word-form + lemma) that the MSD is given its semantics, 
\ie this makes it possible to exemplify how a MSD is used in practice.

The sizes of the \mte\ lexicons vary considerably between the languages: 
the Slovak and Macedonian ones, with around 80,000 lemmas, are quite comprehensive, 
the majority offer medium sized lexicons in the range of 15--50,000 lemmas, and a few are smaller, with Persian only covering the lemmas of ``1984'' and Resian simply giving examples for each MSD.
However, even the smaller lexicons cover the most morphologically complex words, 
such as pronouns and high frequency open class words, providing a good starting point for the development of more extensive lexical resources. 

\section{The ``1984'' corpus}
\label{sec:1984}

The parallel \mte\ corpus consists of the novel “1984” by G. Orwell and its translations. 
The complete novel has about 100,000 tokens, although this of course differs from language to language.  
This corpus is small, esp.\ by today’s standards, and consists of only one text; 
nevertheless, it provides an interesting experimentation dataset, as there are still very few uniformly annotated many-way parallel corpora.

The corpus is available in a format where the novel is extensively annotated for structures which would be mostly useful in the context of a digital library, such as sections, paragraphs, verse, lists, notes, names, etc. More interestingly, the “1984” also exists as a separate corpus, which uses only basic structural tags but 
annotates each word with its context-disambiguated and – for the most part – hand-validated MSD and lemma. 
This dataset provides the final piece of the morphosyntactic triad, as it
contextually validates the specifications and lexicon and provides
examples of actual usage of the MSDs and lexical items. 
It is useful for training part-of-speech taggers and lemmatisers, or for studies involving word-level syntactic information in a multilingual setting, 
such as factored models of statistical machine translation.

\subsection{The linguistically annotated corpus}

As illustrated in \reffig{fig:1984}, the text body consists of basic structure (divisions, paragraphs, sentences) 
and the tokenised text, where the words are annotated by (a pointer to) their MSD and the lemma. 
The elements and attributes for the linguistic annotation come from the TEI analysis module.
The document also contains, in its back matter, the feature and feature-value libraries, automatically derived from the language specific morphosyntactic specifications. 
The feature-value library defines the MSDs, by giving them identifiers and decomposing them into features, \ie giving pointers to their definitions, while
the feature library provides these definitions in the form of attribute-value pairs. 
Each linguistically annotated “1984” thus contains within it the mapping from the MSD tags to the equivalent feature structures. 

\begin{figure}[ht]
\centering
\begin{verbatim}
<text xml:id="Osl." xml:lang="sl">
 <body>
  <div type="part" xml:id="Osl.1">
   <div type="chapter" xml:id="Osl.1.2">
     <p xml:id="Osl.1.2.2">
      <s xml:id="Osl.1.2.2.1">
       <w lemma="biti" ana="#Va-p-sm">Bil</w>
       <w lemma="biti" ana="#Va-r3s-n">je</w>
       <w lemma="jasen" ana="#Agpmsnn">jasen</w>
       <c>,</c>
       <w lemma="mrzel" ana="#Agpmsnn">mrzel</w>
       <w lemma="aprilski" ana="#Agpmsny">aprilski</w>
       <w lemma="dan" ana="#Ncmsn">dan</w>
       ...
 </body>
 <back>
  ...
   <fLib>
    <f name="CATEGORY" xml:id="N0." xml:lang="en">
     <symbol value="Noun"/>
    </f>
    <f name="Type" xml:id="N1.c" xml:lang="en">
     <symbol value="common"/>
    </f>
    <f name="Type" xml:id="N1.p" xml:lang="en">
     <symbol value="proper"/>
    </f>
    <f name="Gender" xml:id="N2.m" xml:lang="en">
     <symbol value="masculine"/>
    </f>
     ...
   </fLib>
   <fvLib>
    <fs xml:id="Ncmsn" xml:lang="en" 
         feats="#N0. #N1.c #N2.m #N3.s #N4.n"/> 
    <fs xml:id="Ncmsg" xml:lang="en" 
         feats="#N0. #N1.c #N2.m #N3.s #N4.g"/>>
     ...
   </fvLib>
  </back>
 </text>
\end{verbatim}
\caption{Linguistically encoded ``1984'' with feature and feature structure definitions.}
\label{fig:1984}
\end{figure}

To further illustrate the annotation we give in Appendix 1 the first sentence of ``1984'' for all the languages that have this annotated corpus in \mte.

\subsection{Sentence alignments}

The ``1984'' corpus also comes with separate files containing
sentence alignments between the languages. 
In addition to the hand-validated alignments between English and the translations
Version 4 also includes automatically induced pair-wise alignments between all the languages, as well as a multi-way alignment spanning all the languages.
The problem of producing optimal n-way alignments from (high-precision) 2-way alignments with a hub is interesting, and more complex than might be obvious at 
first sight, as the source alignments need not be 1:1, and the alignment of different languages can have different spans of such $m:n$ alignments ($m, n \geq 0$);
the Java program used to compute them \citep{cerep:aligner} is available from the download page of \mte.

\reffig{fig:alg} shows a few sentence links from the two-way alignment between Macedonian and Slovene. 
Each link gives the arity of the alignment and a series of (sentence) targets. The \code{@target} attribute is
in TEI  defined as a series of 2 or more values of XML schema type \code{anyURI}, 
so a target must be (unlike CES) fully qualified and it is not possible to directly distinguish between the two languages of the alignment.
However, this is easily done via the value of the \code{@n} attribute or by the \code{@xml:lang} attribute of the 
(ancestor of) the referred to sentences. 
Given that there have to be two or more URIs as the value 
of \code{@targets} it is also not possible to encode 1:0 and 0:1 alignments which in CES used to be encoded explicitly.
Whether the lack of such alignment ever makes a difference in practice, is an open question.

\begin{figure}
\centering
\begin{verbatim}
<linkGrp type="alignment" corresp="oana-mk.xml oana-sl.xml">
  <link n="1:1" 
   targets="oana-mk.xml#Omk.1.1.1.1 oana-sl.xml#Osl.1.2.2.1"/>
  <link n="1:1" 
   targets="oana-mk.xml#Omk.1.1.1.2 oana-sl.xml#Osl.1.2.2.2"/>
   ...
  <link n="2:1" 
   targets="oana-mk.xml#Omk.1.1.2.6 oana-mk.xml#Omk.1.1.2.7 
            oana-sl.xml#Osl.1.2.3.6"/>
  <link n="1:2" 
   targets="oana-mk.xml#Omk.1.1.2.8 oana-sl.xml#Osl.1.2.3.7 
            oana-sl.xml#Osl.1.2.3.8"/>
  ...
  <link n="1:1" 
    targets="oana-mk.xml#Omk.4.23.6 oana-sl.xml#Osl.4.25.7"/>
  <!--link n="0:1" targets="oana-sl.xml#Osl.4.12.2"/-->
</linkGrp>

\end{verbatim}
\caption{Example of sentence alignments for ``1984''.}
\label{fig:alg}
\end{figure}

\section{Related work}
\label{sec:related}

This section reviews work which connects to the \mte\ resources, \ie making available multilingual morphosyntactic specifications, lexicons and annotated parallel corpora. 

\subsection{Morphosyntactic specifications}

Harmonisation of multilingual linguistic features usually proceeds in the scope of large international projects, 
and while \mte\ has its genesis in such efforts, in particular EAGLES and \mt, it has since proceeded by slowly adding new languages and updating the encoding of the resources, without making any revolutionary changes to the basic concept.
In the meantime other initiatives have also been cataloguing and standardising linguistic features, although on a much broader scale, not limited to morphosyntax. 

GOLD, the General Ontology for Linguistic Description \citep{glot-gold} is an effort to
create a freely available domain-specific ontology for linguistic concepts. 
This is a well advanced effort, where (morphosyntactic) terms are extensively documented, also with references to literature.
As the complete ontology is also available for download, it would be interesting to link the categories, attributes and their values form the \mte\ specifications to GOLD, thus providing an explication of their semantics. 

Mostly as a result of a series of EU projects, a number of standards for encoding linguistic resources have been (or are being) developed 
by the ISO Technical Committee TC 37 ``Terminology and other language and content resources'', in particular its Subcommittee SC 4 ``Language resource management''. 
Morphosyntactic features are, along with other linguistic features, defined in the ISO standard 12620:2009 
``Specification of data categories and management of a Data Category Registry for language resources'', and
the standard is operationalized as the isoCat Web service at \textit{http://www.isocat.org/}.

The isoCat Data Category Registry (DCR) \citep{lrec08:isocat,FrDeSoDeMo2008} assigns PIDs (permanent indentifiers) to data categories, such as morphosyntactic features, and these PIDs then serve as stable identifiers for particular features. Users can also browse or search for data categories, export a chosen subset, or add new categories. 
The GOLD ontology has also been added to isoCat, although the information accompanying the features is not given in isoCat; rather the data categories just refer to the GOLD site.
While \mte\ has served as one of the sources for developing the ISO DCR, it has not been so far directly included in isoCat as one of the possible profiles. 
This would certainly be a useful endeavour but is complicated by the fact that, unlike GOLD, the DCR registry is not available for download and upload, 
which precludes (semi)automatically adding already existing category registries.
There are currently also some technical and conceptual problems with adding existing feature collections, 
as documented for the case of mapping the National Corpus of Polish tagset to ISOcat \citep{prz:isocat}.

GOLD and isoCat deal with linguistic features and do not propose specific multilingual harmonised tagsets. 
Surprisingly, it is only relatively recently that research has moved in this direction. 
After \mt\ and \mte\ probably the first and very partial attempt in this direction was the dataset used in the 
CoNLL-X shared task on Multilingual Dependency Parsing \citep{Buchholz:2006:CST:1596276.1596305}, which consisted of 
uniformly encoded treebanks for 13 languages. However, while the format of the treebanks was the same, there was no attempt to unify the PoS tagsets or morphosyntactic features of the treebanks.

A more interesting approach is that taken in Interset 
\citep{biblio-6437616343801484763,biblio6014001168528455610}, even though it does not propose multilingual tagsets.
Rather, the idea is to introduce a central and largely universal set of features and then write drivers from and to particular tagsets to this pivot feature set. 
Then, if a particular tagset A needs to be made compatible with another tagset B (either for the same or for another language) it is enough to run the driver for A into the pivot and the driver for B from the pivot. 
So, for each tagset only two drivers need to be specified, enabling the conversion to and from all the covered tagsets.
There are of course quite a few problems in defining such mappings, such as partial overlap of features, but the approach has been validated in practice and the problems and suggested solutions are discussed in the literature. The Interset approach has been subsequently also used for tagset harmonisation of treebanks for 29 languages, which, together with the harmonisation of syntactic dependencies, resulted in the 
HamleDT, the Harmonized Multi-Language Dependency Treebank 
\citep{hamledt}.

The most influential multilingual tagset is probably the ``Universal tagset'' proposed by \cite{PETROV12.274}, which maps tagsets for 22 languages to a tagset consisting of 12 different tags. 
While such a tagset is undoubtedly useful, it does propose only a lowest common denominator for the languages, thus losing most information from the original tagsets.

\subsection{Morphosyntactic lexicons}

\mte\ resources also offer morphosyntactic lexicons for languages for which they are otherwise still hard to obtain. 
ELDA, for example, offers almost 600 lexicons, but most are for Western European languages, and are, for the most part, not for free. 
LDC, on the other hand, offers cheaper resources but has very few lexicons, and those mostly for speech or for very low resourced languages. 
It should be noted, however, that ELDA does offer the lexicons of the MULTEXT project, \ie 
for English, French, German, Italian, Spanish, and Castilian, which complement the \mte\ lexicons.

\subsection{Parallel annotated corpora}

Finally, the \mte\ parallel ``1984'' corpus is, of course, very small and too uniform to seriously train taggers but, again, available parallel tagged and hand validated corpora are quite difficult to find, so it represents a viable option for developing tagger and lemmatiser models. 
The text is also interesting from a literary and translation perspective: 
the novel ``1984'' is an important work of fiction and linguistically quite interesting, \eg 
in the choices the translators made in translating Newspeak words into their language. 
Again, ELDA does offer (a part of) the MULTEXT corpus, which contains passages from the Written Questions and Answers of the Official Journal of the European Community, with the same languages as for the lexicons. However, only English, French, Italian and Spanish parts are tagged, with roughly 200,000 words per language. 

Many other highly multilingual corpora have, of course, also become available in the many years since MULTEXT, with the best known being Europarl \citep{koehn2005epc}, JRC-ACQUIS \citep{DBLP:journals/corr/abs-cs-0609058} and other corpora compiled by JRC. But while these corpora contain 20+ languages and are quite large, the texts are not word-level (PoS tags, lemmas) annotated and available corpora with such annotations continue to be rather rare.

\section{Conclusions}
\label{sec:conc}

This chapter has introduced one of the oldest maintained sets of multilingual resources, covering mostly the morphosyntactic level of linguistic description. From the beginning, the objective of \mte\ has also been to make its resources available to the wider research community. While it proved impossible to distribute the resources in a completely open manner, a portion of the resources is freely available for download or browsing and for the rest, the user has to agree to use them for research only and to acknowledge their use, and is then free to download them from the project Web site. 

Further work on the resources could proceed in a number of directions. As mentioned, an obvious next step in the development of the specifications and associated tagsets would be to link them to universal vocabularies, in particular isoCat and GOLD. 

The second direction concerns the quality of the resources: it has been noted that the \mte\ morphosyntactic specifications lack consistency between the languages \citep{prz:wol:03p,mte:ukr,feldman:light}. 
Specific points are summarised in \cite{rosen:tags} and can be divided into cases where different features in various languages are used to describe the same phenomenon, and, conversely,  the same features are used to describe different phenomena, and that certain features are too specific and hard to extend to cover similar phenomena in another language; in short, the harmonisation of the specifications between the languages is less than perfect. 
There are several reasons for this, most already mentioned: 
the specifications typically reflect the annotations in some source lexicon for the language, and the logic of such language and resource particular morphosyntactic annotations. 
The linguistic traditions of different languages differ, and this is also reflected in the choice and configuration of the features. 
Some steps in harmonising the \mte\ specifications have already been undertaken in the context of converting them into an OWL DL ontology \citep{chiracos:mte}, which enables logical inferences over feature sets to be made on the basis of partial information. This process also pin-pointed inconsistencies, which were, to an extent, resolved in the ontology. 

The specifications also provide a framework in which other, different morphosyntactic tagsets can be defined. For Slovene, we have used the framework to define two new sets of morphosyntactic specifications with associated tagsets. The SPOOK corpus \citep{Erjavec:SPOOK} is a corpus of parallel sentence aligned bi-texts, where one of the languages is Slovene, with the other being English, German, French or Italian. The SPOOK foreign language texts have been tagged with TreeTagger \citep{schmid1994} which is 
distributed with a language models covering the SPOOK foreign languages, but having very disparate tagsets. 
To harmonise these tagsets, we developed the SPOOK morphosyntactic specifications, where the TreeTagger tags are 1:1 mapped onto MSDs for each particular language, using,
where necessary, idiosyncratic features. With this it is possible to use the corpora either with the source TreeTagger tags or with harmonised SPOOK MSDs.
The other case concerns corpora of historical Slovene \citep{ERJAVEC12.445}, where the focus of the project was on modernisation of historical word-forms, rather than on MSD tagging. Nevertheless, we also wanted to annotate at least basic PoS information on the words. To this end, we developed the IMP morphosyntactic specifications, based on the \mte\ ones for Slovene, which, however, strip all inflectional features from the tags, resulting in a small tagset of 32 MSDs. Both specifications are available on-line, in the same formats as the \mte\ ones.


\bibliographystyle{spbasic}      

\section*{Appendix 1. Examples of annotated text from Orwell's ``1984''.}
\includegraphics[width=1.0\textwidth]{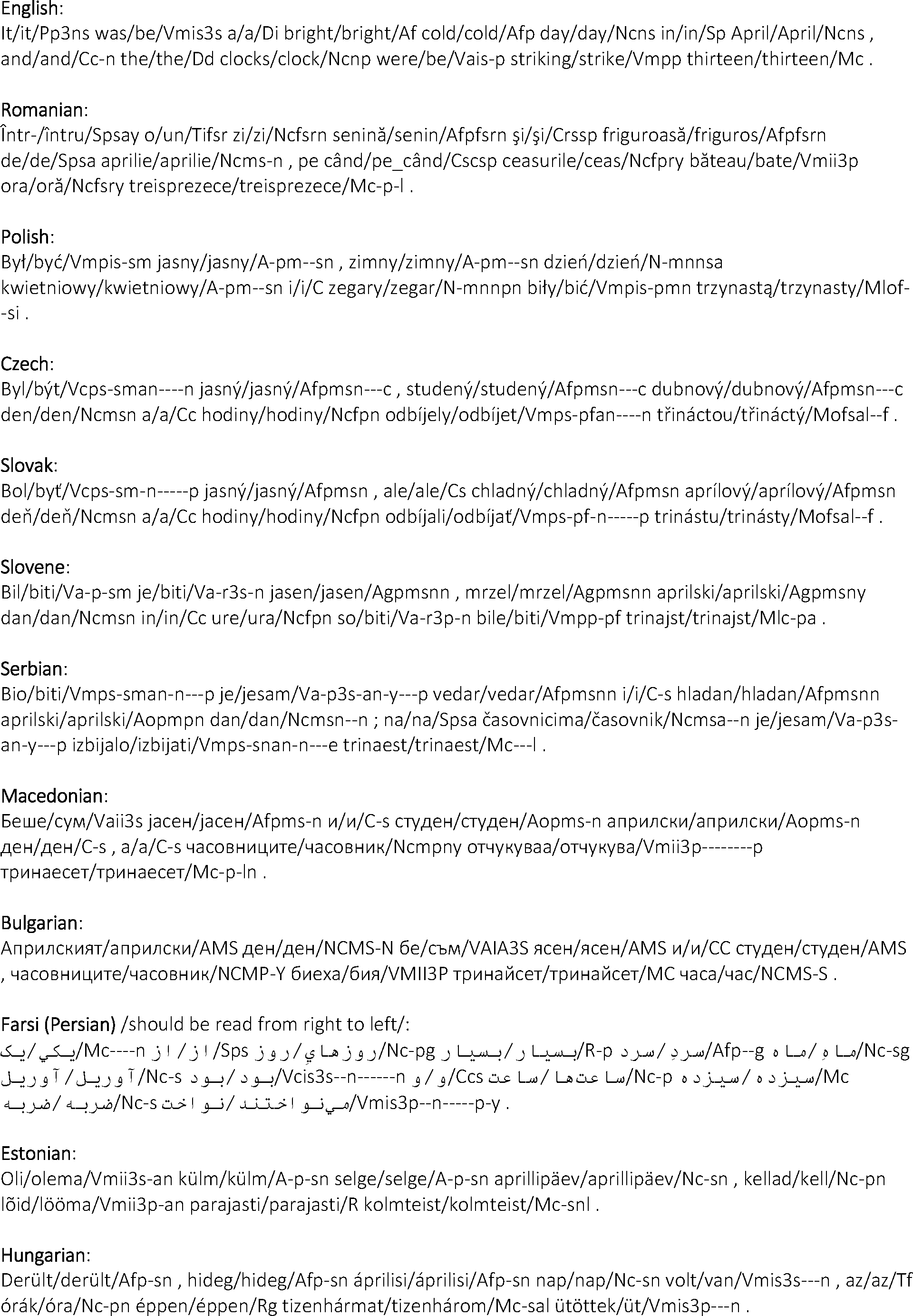}

\end{document}